\documentclass{article}

\usepackage{arxiv}

\usepackage{amsmath}

\usepackage[utf8]{inputenc} 
\usepackage[T1]{fontenc}    
\usepackage{hyperref}       
\usepackage{url}            
\usepackage{booktabs}       
\usepackage{amsfonts}       
\usepackage{nicefrac}       
\usepackage{microtype}      
\usepackage{cleveref}       
\usepackage{graphicx}
\usepackage[round]{natbib}
\usepackage{doi}

\usepackage{tcolorbox}
\usepackage{tabularx}
\usepackage{circuitikz}
\usepackage{tikz}
\usetikzlibrary{fadings}
\usepackage{multirow}
\usepackage{algorithm}
\usepackage{algorithmic}

\definecolor{ubbl}{cmyk}{0.09, 0.04, 0, 0.02}

\newcommand{\bigconc}{\mathop{\mathpalette\bigconcinn\relax}}
\newcommand{\bigconcinn}[2]{%
  \vcenter{\hbox{$\bigconcchoose#1\bigconcsize|\mkern1mu\bigconcsize|$}}}
\newcommand{\bigconcchoose}[1]{\def\bigconcsize{}%
  \ifx#1\displaystyle
    \let\bigconcsize\Big
  \else
    \ifx#1\textstyle
      \let\bigconcsize\big
    \fi
  \fi#1}

\title{Informed Named Entity Recognition Decoding for Generative Language Models}

\date{}

\usepackage{authblk}

\newbox{\orcid}\sbox{\orcid}{\includegraphics[scale=0.06]{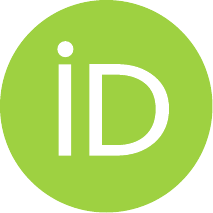}} 
\author[1,2]{%
	\href{https://orcid.org/0000-0003-4685-0847}{\usebox{\orcid}\hspace{1mm}Tobias Deu{\ss}er\thanks{\texttt{tdeusser@uni-bonn.de}}}%
}
\author[2]{%
	\href{https://orcid.org/0000-0002-5496-4177}{\usebox{\orcid}\hspace{1mm}Lars Hillebrand}
}
\author[1,2]{%
	\href{https://orcid.org/0000-0001-6615-2128}{\usebox{\orcid}\hspace{1mm}Christian Bauckhage}
}
\author[1,2]{%
	\href{https://orcid.org/0009-0004-6680-8210}{\usebox{\orcid}\hspace{1mm}Rafet Sifa}
}
\affil[1]{University of Bonn, Bonn, Germany}
\affil[2]{Fraunhofer IAIS, Sankt Augustin, Germany}

\renewcommand{\shorttitle}{Informed Named Entity Recognition Decoding for Generative Language Models}

\hypersetup{
pdftitle={Informed Named Entity Recognition Decoding for Generative Language Models},
pdfsubject={cs.CL, cs.AI},
pdfauthor={Tobias Deu{\ss}er, Lars Hillebrand, Christian Bauckhage, Rafet Sifa},
pdfkeywords={Named Entity Recognition, Information Extraction, Large Language Models, Natural Language Processing, Machine Learning},
}

\begin{document}
\maketitle
\addtocounter{footnote}{-1}
\begin{abstract}
Ever-larger language models with ever-increasing capabilities are by now well-established text processing tools. Alas, information extraction tasks such as named entity recognition are still largely unaffected by this progress as they are primarily based on the previous generation of encoder-only transformer models.  Here, we propose a simple yet effective approach, Informed Named Entity Recognition Decoding (iNERD), which treats named entity recognition as a generative process. It leverages the language understanding capabilities of recent generative models in a future-proof manner and employs an informed decoding scheme incorporating the restricted nature of information extraction into open-ended text generation, improving performance and eliminating any risk of hallucinations. We coarse-tune our model on a merged named entity corpus to strengthen its performance, evaluate five generative language models on eight named entity recognition datasets, and achieve remarkable results, especially in an environment with an unknown entity class set, demonstrating the adaptability of the approach.
\end{abstract}

\keywords{Named Entity Recognition \and Information Extraction \and Large Language Models \and Natural Language Processing \and Machine Learning}

\section{Introduction}

Recent public releases of large language models (LLMs) with human-like writing skills have drawn unprecedented attention to natural language processing (NLP). Indeed, the performance of transformer-based LLMs increases notably, and they develop ``emergent abilities'', i.e. their performance increases significantly, when their number of parameters exceeds a certain level \citep{wei2022emergent}. 

On the other hand, tasks not based on generative transformers, say sentiment analysis, contradiction detection, or named entity recognition, have been relegated to the backseat of this latest push in NLP. As of this writing, they are usually tackled using ``encoder-only''\footnote{``Encoder-only'' refers to transformer models which only consist of encoder blocks. This contrasts with the original encoder-decoder structure proposed by \citet{vaswani2017attention} or the ``decoder-only'' structure of generative models.} language models \citep{heinsen2022algorithm,deusser2023contradiction,verma-etal-2023-comparing} which are typically much smaller than their ``decoder-only'' counterparts. 

Here, we intend to narrow the gap between generative and extractive NLP and introduce a novel named entity recognition (NER) framework. Our Informed Named Entity Recognition Decoding (iNERD) approach has three main features: First, it leverages proven capabilities of ``decoder-only'' models. Our current approach works with the latest generative LLMs but can easily incorporate even better models once they become available and thus keep up with rapid release cycles \citep{zhao2023survey} making it future-proof and quick to upgrade. 

Second, we exploit the extensive pre-training and the resulting language understanding capabilities of state-of-the-art LLMs. Our approach involves an informed decoding algorithm which eliminates any hallucinations current models might suffer from during our approach \citep{bang2023multitask} and improves performance by ruling out impossible tokens during generation. To strengthen the model's understanding of the NER task, we ``coarse-tune'' it on a merged corpus of various task-specific datasets.

Third, we propose a simple decoding strategy which allows for casting the extractive task of named entity recognition as a generative task. Our idea is to let the model generate extended texts of the following form:

\begin{tcolorbox}[notitle,boxrule=0pt,
boxsep=0pt,left=0.6em,right=0.6em,top=0.5em,bottom=0.5em,
colback=gray!10,
colframe=gray!10]
``EU rejects German call to boycott British lamb. \textbf{\textless CT\textgreater} Organisation \textbf{\textless TCS\textgreater} EU \textbf{\textless ES\textgreater} Location \textbf{\textless TCS\textgreater} German \textbf{\textless ES\textgreater} Location \textbf{\textless TCS\textgreater} British \textbf{\textless ES\textgreater}'',
\end{tcolorbox}

\noindent Here, the special tokens inside angular brackets signal the start of the entity string (\textbf{\textless C}ombine\textbf{T}oken\textbf{\textgreater}), separate entity type and entity content (\textbf{\textless T}ype\textbf{C}ontent\textbf{S}eparator\textbf{\textgreater}), and identify different entities (\textbf{\textless E}ntity\textbf{S}eparator\textbf{\textgreater}). During inference, we enforce this structure and thus reduce the complexity of the generation step. 

Extensive evaluations show that this approach achieves remarkable performances in various NER settings ranging from general-purpose over bio-medical to finance. 

In short, our contributions presented in this paper are the following:

\begin{itemize}
    \item We propose a novel future-proof architecture to cast the extractive process of named entity recognition as a generative one, incorporating natural language understanding capabilities of generative models into the process.
    \item We introduce a novel decoding strategy for such an architecture, which prevents the model from hallucinating and improves performance.
    \item We ``coarse-tune'' decoder-only models like Llama \citep{touvron2023llama} or GPT-2 \citep{radford2019language} on a merged named entity recognition dataset to further improve the contextual awareness of these models for NER tasks. 
    \item We publicly provide our code as well as the weights of our best-performing model\footnote{The link to the GitHub repository will be published upon acceptance of this paper.}.
\end{itemize}

Next, we review recent related work on named entity recognition and generative language models. We then elaborate on our framework, our encoding scheme for named entities, and the corresponding informed decoding. Afterwards, we discuss our experimental protocol and present and discuss the results obtained on eight benchmark  datasets. Finally, we summarize our main results and provide an outlook to auspicious future work.

\section{Related Work}

Named entity recognition \citep{grishman-sundheim-1996-message} is a fundamental task in text mining and natural language processing. Among others, it allows for anonymization \citep{10.1162/coli_a_00458} or relation extraction \citep{kpibert} and, owing to its practical importance, has been studied w.r.t.~standardized corpora early on (e.g. the CoNLL-2003 data collected by \citet{tjong-kim-sang-de-meulder-2003-introduction}).

Prior to the deep learning revolution, NER was usually tackled in a rule-based manner  
(\citealp{ETZIONI200591}) or with unsupervised- or feature-based supervised learning (\citealp{Collins1999UnsupervisedMF,Zhang2013UnsupervisedBN,bikel-etal-1997-nymble,mcnamee-mayfield-2002-entity}).

In their seminal paper on BERT, an encoder-only transformer, \citet{devlin2018bert} achieved remarkable results on the CoNLL-2003 data by adding a classifier on top of the encoder and fine-tuning the model. Much subsequent work on similar approaches towards NER then focused on improved context awareness. To name but a few, \citet{Luo2019HierarchicalCR} fused hierarchical contextualized representations with input token embeddings, \citet{biobert} applied additional pre-training aimed at biomedical texts, and \citet{wang-etal-2021-improving} added a conditional random field on top of BERT. Going even further, \citet{yamada-etal-2020-luke} forced entity extraction during pre-training and \citet{zhou-chen-2021-learning} added a co-regularization framework for entity-centric information extraction, to achieve state-of-the-art results. Nevertheless, all of these approaches are built upon an encoder-only transformer model and are unsuited to incorporate the decoder-only transformer architecture powering the recent popularity and success of natural language processing.

Closest to the ideas proposed in this paper, \citet{yan-etal-2021-unified-generative} formulated NER as an entity span sequence generation task, in which they added special tokens to their vocabulary to then generate entities and their types in an autoregressive fashion. \citet{NEURIPS2022_63943ee9} extended this to cover more tasks in the information extraction field. The advantage of our approach is that we do \textit{not} add the entity type tokens as special tokens, but as regular tokens already known to the model. Furthermore, \citet{wang2023gptner} leveraged the GPT--3 \citep{gpt3} API to tag entities in a sentence in a zero and few-shot approach.

Generative language models gained widespread public interest with the introduction of GPT--3 \citep{gpt3} and GPT--4 \citep{openai2023gpt4}, which both reported impressive language understanding and writing capabilities, but did not make their models and exact architectures known to the research community. On the other hand, their predecessors, GPT--2 \citep{radford2019language} and GPT \citep{radford2018improving}, are openly available and were the first to implement a ``decoder-only'' architecture, which discarded the Encoder-Decoder structure proposed in \citet{vaswani2017attention} in favour of an autoregressive generation process, trained by teacher-forcing.

In recent years, this field expanded rapidly, driven by its prominent place in public discourse, and many new models emerged and were studied, e.g. Llama \citep{touvron2023llama} and its second iteration \citep{llama2}, RedPajama \citep{together2023redpajama}, Falcon \citep{falcon}, Bloom \citep{workshop2023bloom}, or OPT \citep{zhang2022opt}. As already mentioned in the previous section, a critical property of such a large language model (LLM) is that performance experiences a remarkable increase once the model scale, i.e. its parameter size, surpasses a certain threshold, dubbed ``emergent abilities of LLMs'', studied in \citet{wei2022emergent} and \citet{rae2022scaling}. Due to the sheer size of these models, reaching into the hundreds of billions, it is apparent that training and even fine-tuning them is costly and time-consuming. To alleviate this and make the training of pre-trained LLMs accessible to a broader audience, \citet{hu2021lora} introduced LoRA, a framework that freezes the pre-trained model weights and injects trainable rank decomposition matrices into the transformer layers.

Regardless, these LLMs are trained to be capable text generation tools and are, at their current state, mostly incompatible with other NLP tasks like information extraction, a flaw which we alleviate with the iNERD approach introduced in this work.

\section{Methodology}

Here, we describe how we formulate named entity recognition (NER) as a task suited for generative language models. Following this, we shed light on how our algorithm for Informed Named Entity Recognition Decoding (iNERD) and the complete setup is defined and point out the advantages compared to other approaches. 

\subsection{Named entity decoding} 

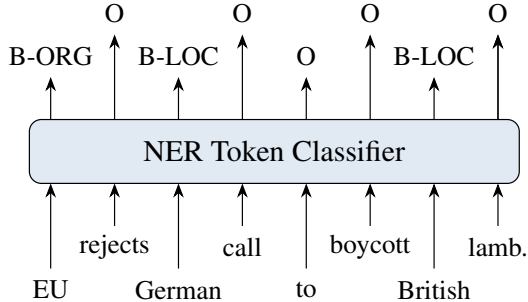
\begin{figure}[!t]
\centering
\resizebox{0.44\columnwidth}{!}{%
\begin{tikzpicture}

\node [font=\footnotesize] at (0.25,14) {EU};
\node [font=\footnotesize] at (1,14.5) {rejects};
\node [font=\footnotesize] at (1.75,14) {German};
\node [font=\footnotesize] at (2.5,14.5) {call};
\node [font=\footnotesize] at (3.25,14) {to};
\node [font=\footnotesize] at (4,14.5) {boycott};
\node [font=\footnotesize] at (4.75,14) {British};
\node [font=\footnotesize] at (5.5,14.5) {lamb.};
\draw [ -Stealth] (0.25,14.25) -- (0.25,15.25);
\draw [ -Stealth] (1,14.75) -- (1,15.25);
\draw [ -Stealth] (1.75,14.25) -- (1.75,15.25);
\draw [ -Stealth] (2.5,14.75) -- (2.5,15.25);
\draw [ -Stealth] (3.25,14.25) -- (3.25,15.25);
\draw [ -Stealth] (4,14.75) -- (4,15.25);
\draw [ -Stealth] (4.75,14.25) -- (4.75,15.25);
\draw [ -Stealth] (5.5,14.75) -- (5.5,15.25);
\draw [ fill={ubbl} , rounded corners, ] (0,16) rectangle node {NER Token Classifier} (5.75,15.25);
\draw [ -Stealth] (0.25,16) -- (0.25,16.5);
\draw [ -Stealth] (1,16) -- (1,17);
\draw [ -Stealth] (1.75,16) -- (1.75,16.5);
\draw [ -Stealth] (2.5,16) -- (2.5,17);
\draw [ -Stealth] (3.25,16) -- (3.25,16.5);
\draw [ -Stealth] (4,16) -- (4,17);
\draw [ -Stealth] (4.75,16) -- (4.75,16.5);
\draw [ -Stealth] (5.5,16) -- (5.5,17);
\node [font=\footnotesize] at (0.25,16.75) {B-ORG};
\node [font=\footnotesize] at (1,17.25) {O};
\node [font=\footnotesize] at (1.75,16.75) {B-LOC};
\node [font=\footnotesize] at (2.5,17.25) {O};
\node [font=\footnotesize] at (3.25,16.75) {O};
\node [font=\footnotesize] at (4,17.25) {O};
\node [font=\footnotesize] at (4.75,16.75) {B-LOC};
\node [font=\footnotesize] at (5.5,17.25) {O};
\draw [ -Stealth] (5.5,16) -- (5.5,17);
\end{tikzpicture}
}%
\caption{Illustration of Named Entity Recognition as a token classification task with the IOB tagging scheme. Each input token is classified either as B-Entity type, I-Entity type, or O.}
\label{fig:ner_token_classifier}
\end{figure}

NER is usually formulated as a ``token classification'' task, as seen in \citet{Dou2023DomainAdaptedDP} or \citet{Nguyen2023AUCMF}. In such a setup, an embedding of each token is generated using a text encoder, often an encoder-only transformer model like BERT \citep{devlin2018bert}. This embedding is then fed into a classifier, which can be anything from a simple logistic regression to a more involved deep neural network to classify each token as either a part of an entity or not. This prediction generally has to include the entity start and entity end information, which can be achieved with, among others, the \textit{IOB} tagging scheme \citep{ramshaw-marcus-1995-text}. Figure~\ref{fig:ner_token_classifier} illustrates this setup.

In contrast, we propose to model this task as a generative process, simplifying its machine-learning components to just one building block: a decoder-only transformer model. 

To formalize this, we define the input $I$ for our generative model during the training phase for $n$ entities $e$ as

\begin{align}
\begin{split}
I &= I_s \oplus \kappa \oplus \bigconc^n \left( \xi_e \oplus \tau \oplus I_e \oplus \epsilon \right) \label{eq:nerd}\\
 &= I_s \oplus \kappa \oplus E,
\end{split}
\end{align}

\noindent where $\oplus$ is the concatenation operator, $I_s$ the actual sentence from which we intend to extract entities, $\kappa$ the ``combine'' token, $\xi_e$ the type of entity $e$, $\tau$ the ``type-content'' separator token, $I_e$ the actual entity string, $\epsilon$ the ``entity separator'' token, and $\bigconc^n$ concatenates its input along the number of entities $n$. This concatenation $\bigconc^n\left( \xi_e \oplus \tau \oplus I_e \oplus \epsilon \right)$ is the entity string $E$ of our input $I$, i.e. what is unknown during inference and has to be predicted.

To make Equation~\ref{eq:nerd} more accessible, we can review the example from the introduction,

\begin{tcolorbox}[notitle,boxrule=0pt,
boxsep=0pt,left=0.6em,right=0.6em,top=0.5em,bottom=0.5em,
colback=gray!10,
colframe=gray!10]
``EU rejects German call to boycott British lamb. \textbf{\textless CT\textgreater} Organisation \textbf{\textless TCS\textgreater} EU \textbf{\textless ES\textgreater} Location \textbf{\textless TCS\textgreater} German \textbf{\textless ES\textgreater} Location \textbf{\textless TCS\textgreater} British \textbf{\textless ES\textgreater}'',
\end{tcolorbox}

\noindent in which 

\begin{itemize}
    \item $I_s$ is the input sentence \textit{``EU rejects German call to boycott British lamb''},
    \item $\kappa$ the string ``\textit{\textless CT\textgreater}'',
    \item $\xi_e$ the entity types \textit{``Organisation''} and \textit{``Location''},
    \item $\tau$ the string ``\textit{\textless TCS\textgreater}'',
    \item $I_e$ the actual entity content \textit{``EU''}, \textit{``German''} and \textit{``British''},
    \item $\epsilon$ the string ``\textit{\textless ES\textgreater}'',
    \item $E$ the entity string ``\textit{Organisation \textit{\textless TCS\textgreater} EU \textit{\textless ES\textgreater} Location \textit{\textless TCS\textgreater} German \textit{\textless ES\textgreater} Location \textit{\textless TCS\textgreater} British \textit{\textless ES\textgreater}}''
\end{itemize}

We can then fine-tune the pre-trained decoder-only model to predict each token of the input $I$ autoregressively using teacher forcing \citep{Williams1989ALA}, i.e. the causal language modelling task is unchanged for these models. We calculate the loss on all predicted tokens after the $\kappa$ token.

Compared to the approach introduced in \citet{yan-etal-2021-unified-generative}, the essential advantage of our framework for named entity decoding is that we do \textit{not} add entity type tokens $\xi$ as special tokens, but as regular tokens already known to the model. Their approach, where the example sentence above becomes ``EU rejects German call to boycott British lamb. \textit{\textless ORG\textgreater} EU \textit{\textless LOC\textgreater} German \textit{\textless LOC\textgreater} British'', loses the meaningful embedding a transformer model has learned for $\xi$, i.e. the model has to learn anew what the introduced special tokens mean. 

\subsection{Informed named entity recognition decoding}

In the previous section on \textit{Named entity decoding}, we only considered the training process, in which we apply teacher forcing to correct the model if it ``makes a mistake'' during the generation to accelerate convergence. However, during inference, applying teacher forcing would either be cheating or simply impossible if no ground truth exists. 

Nevertheless, we do know quite a bit about what tokens to expect at a certain point during inference, described by these four rules:

\begin{enumerate}
    \item After the combine token $\kappa$ or the entity separator token $\epsilon$, the entity type token $\xi$ or the end-of-sequence token has to be predicted.
    \item After predicting the entity type $\xi$, the type-content separator $\tau$ has to be predicted.
    \item After the type-content separator $\tau$, any token from the input $I_s$ may be predicted (signalling the start of the entity $e$).
    \item After a token from the input $I_s$ has been predicted, the only allowed tokens for prediction are either the entity separator token $\epsilon$ (signalling the end of the entity $e$) or the token following the previous token in the input $I_s$ (signalling the continuation of the entity $e$).
\end{enumerate}

These four rules comprise the Informed Named Entity Recognition (iNERD) algorithm, as illustrated in Algorithm~\ref{alg:iNERD}. This algorithm is implemented as a post-processing step and is executed after the model calculates the score over its vocabulary and before mapping this score to the actual token to be predicted. 

\begin{algorithm}[tb]
\caption{iNERD for a batch size of 1}
\label{alg:iNERD}
\textbf{Input}: Scores $S$ with the size of the vocabulary, input IDs $I$ holding the considered sentence and prior predictions\\
\textbf{Parameters}: Combine token $\kappa$, entity separator token $\epsilon$, type content separator token $\tau$, entity type tokens $\xi$\\
\textbf{Output}: Updated scores $S$ with iNERD applied\\
\begin{algorithmic}[1] 
\STATE Let $p$ be the previously predicted token, i.e. the last token in the sequence $I$.
\STATE Let $g$ be a boolean value representing if we are in the ``entity generation phase'', i.e. if in the reversed sequence of $I$ we can find the token $\tau$ before we can find the token $\epsilon$.
\STATE Let $I_s$ be the sentence considered, i.e. everything of $I$ before the token $\kappa$.
\IF {$p = \kappa$ or $p = \epsilon$}
\STATE Mask $S$ to only allow $\xi$ or the end-of-sequence token.
\ELSIF{$p \in \xi$}
\STATE Mask $S$ to only allow $\tau$.
\ELSIF{$g$}
\IF{$p = \tau$}
\STATE Mask $S$ to only allow tokens present in $I_s$.
\ELSE
\STATE Mask $S$ to only allow the token after $p$ in $I_s$ or $\epsilon$.
\ENDIF
\ENDIF
\STATE \textbf{return} $S$
\end{algorithmic}
\end{algorithm}

The advantages of this approach are clear: First, the decoder-only model is unable to hallucinate, as any prediction that does not follow the decoding scheme introduced in Equation~\ref{eq:nerd}, is simply masked out, i.e. the score of this token is set to 0. Second, we can apply this model to unseen data and still expect reasonable results if we define our set of entity type tokens $\xi$ beforehand, as later shown in the Experiments section.

\subsection{Complete Model Setup}

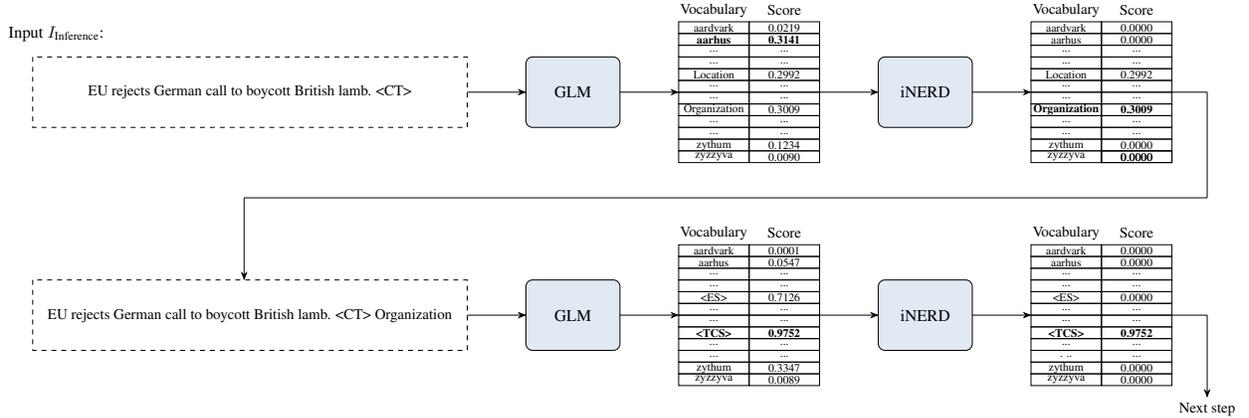
\begin{figure*}[t]
\centering
\resizebox{\textwidth}{!}{%

\begin{tikzpicture}
\tikzstyle{every node}=[font=\normalsize]
\node [font=\normalsize] at (-4,16.75) {Input $I_\text{Inference}$:};
\draw [ -Stealth] (8,15.5) -- (9.25,15.5);
\draw [ -Stealth] (4.75,15.5) -- (6,15.5);

\node [font=\small] at (10,17.25) {Vocabulary};
\node [font=\small] at (11.5,17.25) {Score};
\draw (9.25,15.25) rectangle node {\scriptsize Organization} (10.75,15);
\draw (9.25,15.25) rectangle node {\scriptsize ...} (10.75,15.5);
\draw (9.25,15.5) rectangle node {\scriptsize ...} (10.75,15.75);
\draw (9.25,15.75) rectangle node {\scriptsize Location} (10.75,16);
\draw (9.25,16) rectangle node {\scriptsize ...} (10.75,16.25);
\draw (9.25,16.25) rectangle node {\scriptsize ...} (10.75,16.5);
\draw (9.25,16.5) rectangle node {\scriptsize \textbf{aarhus}} (10.75,16.75);
\draw (9.25,16.75) rectangle node {\scriptsize aardvark} (10.75,17);
\draw (9.25,15) rectangle node {\scriptsize ...} (10.75,14.75);
\draw (9.25,14.75) rectangle node {\scriptsize ...} (10.75,14.5);
\draw (9.25,14.5) rectangle node {\scriptsize zythum} (10.75,14.25);
\draw (9.25,14.25) rectangle node {\scriptsize zyzzyva} (10.75,14);
\draw (10.75,17) rectangle node {\scriptsize 0.0219} (12.25,16.75);
\draw (10.75,16.5) rectangle node {\scriptsize \textbf{0.3141}} (12.25,16.75);
\draw (10.75,16.25) rectangle node {\scriptsize ...} (12.25,16.5);
\draw (10.75,16) rectangle node {\scriptsize ...} (12.25,16.25);
\draw (10.75,15.75) rectangle node {\scriptsize 0.2992} (12.25,16);
\draw (10.75,15.5) rectangle node {\scriptsize ...} (12.25,15.75);
\draw (10.75,15.25) rectangle node {\scriptsize ...} (12.25,15.5);
\draw (10.75,15) rectangle node {\scriptsize 0.3009} (12.25,15.25);
\draw (10.75,14.75) rectangle node {\scriptsize ...} (12.25,15);
\draw (10.75,14.5) rectangle node {\scriptsize ...} (12.25,14.75);
\draw (10.75,14.25) rectangle node {\scriptsize 0.1234} (12.25,14.5);
\draw (10.75,14) rectangle node {\scriptsize 0.0090} (12.25,14.25);
\draw [ -Stealth] (12.25,15.5) -- (13.5,15.5);
\draw [ -Stealth] (15.5,15.5) -- (16.75,15.5);
\node [font=\small] at (17.5,17.25) {Vocabulary};
\node [font=\small] at (19,17.25) {Score};
\draw (16.75,15) rectangle node {\scriptsize \textbf{Organization}} (18.25,15.25);
\draw (16.75,15.25) rectangle node {\scriptsize ...} (18.25,15.5);
\draw (16.75,15.5) rectangle node {\scriptsize ...} (18.25,15.75);
\draw (16.75,15.75) rectangle node {\scriptsize Location} (18.25,16);
\draw (16.75,16) rectangle node {\scriptsize ...} (18.25,16.25);
\draw (16.75,16.25) rectangle node {\scriptsize ...} (18.25,16.5);
\draw (16.75,16.5) rectangle node {\scriptsize aarhus} (18.25,16.75);
\draw (16.75,16.75) rectangle node {\scriptsize aardvark} (18.25,17);
\draw (16.75,15) rectangle node {\scriptsize ...} (18.25,14.75);
\draw (16.75,14.75) rectangle node {\scriptsize ...} (18.25,14.5);
\draw (16.75,14.5) rectangle node {\scriptsize zythum} (18.25,14.25);
\draw (16.75,14.25) rectangle node {\scriptsize zyzzyva} (18.25,14);
\draw (18.25,17) rectangle node {\scriptsize 0.0000} (19.75,16.75);
\draw (18.25,16.75) rectangle node {\scriptsize 0.0000} (19.75,16.5);
\draw (18.25,16.5) rectangle node {\scriptsize ...} (19.75,16.25);
\draw (18.25,16.25) rectangle node {\scriptsize ...} (19.75,16);
\draw (18.25,16) rectangle node {\scriptsize 0.2992} (19.75,15.75);
\draw (18.25,15.75) rectangle node {\scriptsize ...} (19.75,15.5);
\draw (18.25,15.5) rectangle node {\scriptsize ...} (19.75,15.25);
\draw (18.25,15) rectangle node {\scriptsize \textbf{0.3009}} (19.75,15.25);
\draw (18.25,15) rectangle node {\scriptsize ...} (19.75,14.75);
\draw (18.25,14.75) rectangle node {\scriptsize ...} (19.75,14.5);
\draw (18.25,14.5) rectangle node {\scriptsize 0.0000} (19.75,14.25);
\draw (18.25,14.25) rectangle node {\scriptsize 0.0000} (19.75,14);
\draw (18.25,14.25) rectangle node {\scriptsize 0.0000} (19.75,14);
\draw (19.75,15.5) to[short] (20.5,15.5);
\draw (20.5,15.5) to[short] (20.5,13.25);
\draw (20.5,13.25) to[short] (0,13.25);
\draw [ -Stealth] (0,13.25) -- (0,11.5);
\draw [, dashed] (-4.5,16.25) rectangle node {\footnotesize EU rejects German call to boycott British lamb. \textless CT\textgreater} (4.75,14.75);
\draw [ fill={ubbl}, rounded corners, ] (6,16.25) rectangle node {GLM} (8,14.75);
\draw [ fill={ubbl} , rounded corners, ] (13.5,14.75) rectangle node {iNERD} (15.5,16.25);
\draw [, dashed] (-4.5,11.5) rectangle node {\footnotesize EU rejects German call to boycott British lamb. {\textless CT\textgreater}   Organization} (4.75,10);
\draw [ -Stealth] (4.75,10.75) -- (6,10.75);
\draw [ fill={ubbl} , rounded corners, ] (6,11.5) rectangle node {GLM} (8,10);
\draw [ -Stealth] (8,10.75) -- (9.25,10.75);
\draw (9.25,10.5) rectangle node {\scriptsize \textbf{\textless TCS\textgreater}} (10.75,10.25);
\draw (9.25,10.5) rectangle node {\scriptsize ...} (10.75,10.75);
\draw (9.25,10.75) rectangle node {\scriptsize ...} (10.75,11);
\draw (9.25,11) rectangle node {\scriptsize \textless ES\textgreater} (10.75,11.25);
\draw (9.25,11.25) rectangle node {\scriptsize ...} (10.75,11.5);
\draw (9.25,11.5) rectangle node {\scriptsize ...} (10.75,11.75);
\draw (9.25,11.75) rectangle node {\scriptsize aarhus} (10.75,12);
\draw (9.25,12) rectangle node {\scriptsize aardvark} (10.75,12.25);
\draw (9.25,10.25) rectangle node {\scriptsize ...} (10.75,10);
\draw (9.25,10) rectangle node {\scriptsize ...} (10.75,9.75);
\draw (9.25,9.75) rectangle node {\scriptsize zythum} (10.75,9.5);
\draw (9.25,9.5) rectangle node {\scriptsize zyzzyva} (10.75,9.25);
\draw (10.75,12.25) rectangle node {\scriptsize 0.0001} (12.25,12);
\draw (10.75,12) rectangle node {\scriptsize 0.0547} (12.25,11.75);
\draw (10.75,11.75) rectangle node {\scriptsize ...} (12.25,11.5);
\draw (10.75,11.5) rectangle node {\scriptsize ...} (12.25,11.25);
\draw (10.75,11.25) rectangle node {\scriptsize 0.7126} (12.25,11);
\draw (10.75,11) rectangle node {\scriptsize ...} (12.25,10.75);
\draw (10.75,10.75) rectangle node {\scriptsize ...} (12.25,10.5);
\draw (10.75,10.5) rectangle node {\scriptsize \textbf{0.9752}} (12.25,10.25);
\draw (10.75,10.25) rectangle node {\scriptsize ...} (12.25,10);
\draw (10.75,10) rectangle node {\scriptsize ...} (12.25,9.75);
\draw (10.75,9.75) rectangle node {\scriptsize 0.3347} (12.25,9.5);
\draw (10.75,9.5) rectangle node {\scriptsize 0.0089} (12.25,9.25);
\draw [ fill={ubbl} , rounded corners, ] (13.5,11.5) rectangle node {iNERD} (15.5,10);
\draw [ -Stealth] (12.25,10.75) -- (13.5,10.75);
\draw [ -Stealth] (15.5,10.75) -- (16.75,10.75);
\draw (16.75,10.75) rectangle node {\scriptsize ...} (18.25,10.5);
\draw (16.75,11) rectangle node {\scriptsize ...} (18.25,10.75);
\draw (16.75,11) rectangle node {\scriptsize \textless ES\textgreater} (18.25,11.25);
\draw (16.75,11.25) rectangle node {\scriptsize ...} (18.25,11.5);
\draw (16.75,11.5) rectangle node {\scriptsize ...} (18.25,11.75);
\draw (16.75,11.75) rectangle node {\scriptsize aarhus} (18.25,12);
\draw (16.75,12) rectangle node {\scriptsize aardvark} (18.25,12.25);
\draw (16.75,10.5) rectangle node {\scriptsize \textbf{\textless TCS\textgreater}} (18.25,10.25);
\draw (16.75,10.25) rectangle node {\scriptsize ...} (18.25,10);
\draw (16.75,10) rectangle node {\scriptsize. ..} (18.25,9.75);
\draw (16.75,9.75) rectangle node {\scriptsize zythum} (18.25,9.5);
\draw (16.75,9.5) rectangle node {\scriptsize zyzzyva} (18.25,9.25);
\draw (18.25,12.25) rectangle node {\scriptsize 0.0000} (19.75,12);
\draw (18.25,12) rectangle node {\scriptsize 0.0000} (19.75,11.75);
\draw (18.25,11.75) rectangle node {\scriptsize ...} (19.75,11.5);
\draw (18.25,11.5) rectangle node {\scriptsize ...} (19.75,11.25);
\draw (18.25,11.25) rectangle node {\scriptsize 0.0000} (19.75,11);
\draw (18.25,11) rectangle node {\scriptsize ...} (19.75,10.75);
\draw (18.25,10.75) rectangle node {\scriptsize ...} (19.75,10.5);
\draw (18.25,10.5) rectangle node {\scriptsize \textbf{0.9752}} (19.75,10.25);
\draw (18.25,10.25) rectangle node {\scriptsize ...} (19.75,10);
\draw (18.25,10) rectangle node {\scriptsize ...} (19.75,9.75);
\draw (18.25,9.75) rectangle node {\scriptsize 0.0000} (19.75,9.5);
\draw (18.25,9.5) rectangle node {\scriptsize 0.0000} (19.75,9.25);
\draw (19.75,10.75) to[short] (20.5,10.75);
\node [font=\small] at (10,12.5) {Vocabulary};
\node [font=\small] at (11.5,12.5) {Score};
\node [font=\small] at (17.5,12.5) {Vocabulary};
\node [font=\small] at (19,12.5) {Score};
\draw [ -Stealth] (20.5,10.75) -- (20.5,9);
\node [font=\small] at (20.5,8.75) {Next step};

\end{tikzpicture}
}%
\caption{Illustration of the first two steps taken during the inference pipeline of our complete model setup. The model starts with the Input $I_\text{Inference}$ as shown in Equation~\ref{eq:nerd_inference}, which is fed into the generative language model (GLM). This outputs a score vector $S$ over the vocabulary, which in turn is processed by the iNERD algorithm as described in Algorithm~\ref{alg:iNERD}. The highest-scoring token from the vocabulary is then appended to the input, and the process starts anew. This is repeated until the model predicts the end-of-sequence token.}
\label{fig:inerd_in_action}
\end{figure*}

Now, we can build our model with the blocks introduced in the sections on \textit{Named entity decoding} and \textit{iNERD}. First of all, we transform our input to the structure described in Equation \ref{eq:nerd}, which during training contains the entity string $E$, but becomes

\begin{equation}
\label{eq:nerd_inference}
    I_\text{Inference} = I_s \oplus \kappa
\end{equation}

\noindent during inference. This is passed through the generative language model, which assigns a score $s$ to each token in the vocabulary. The resulting score vector $S$ is the input to the iNERD algorithm, as described in Algorithm~\ref{alg:iNERD}. This masks out impossible tokens for the current step, resulting in an updated score vector $S_\text{iNERD}$, which is then used to calculate the next token by taking the one with the highest score $s_\text{iNERD}$. This token is concatenated with the input and the whole process is repeated until the model predicts the end-of-sequence token. Figure~\ref{fig:inerd_in_action} illustrates this procedure for the first two steps.

\section{Experiments}

To practically evaluate the merits of our approach, we conducted experiments on eight datasets. Here, we describe our protocol, discuss data and results, and point out the strengths and flaws. We report the performance in \textit{micro}-F$_1$ in \%, which describes the F$_1$ score, the harmonic mean of precision and recall, when we aggregate the results on all samples individually, independent of the classes. On the other hand, the \textit{macro}-F$_1$ would compute the metric for each class and then average over each of these.

We use the model setup introduced in the previous section for all our experiments and compare it to various benchmarks from other approaches. We test various decoder-only language models for this setup, namely the 1.5 billion parameter GPT2--XL \citep{radford2019language}, the 2.7 billion parameter BioMedLM \citep{biomedlm}, the 3 billion parameter RedPajama \citep{together2023redpajama}, the 7 billion parameter Falcon \citep{falcon}, the 7 and 13 billion parameter Llama \citep{touvron2023llama}, and the 7 billion parameter Llama-2 \citep{llama2}. Additionally, we apply LoRA \citep{hu2021lora}, a framework that freezes the pre-trained model weights while integrating trainable rank decomposition matrices into the transformer layers, to every model with a parameter size above 3 billion.

Our general approach is as follows. We first coarse-tune (see Section ``Coarse-tuning'' for more details)  each language model on our merged NER dataset. We then evaluate each model \textit{without additional fine-tuning} on the test set of each dataset, before we fine-tune them on the respective training set and again report the performance on the test set. Additionally, we conduct an ablation study to highlight the improvements of each component of our approach.

Due to the sheer computational complexity, we only run each experiment once and do not test various seeds to take the average of each run. Furthermore, and for the same reason, we apply no hyperparameter tuning in this scenario. The fixed hyperparameters we used are an (accumulated) batch size of $16$, a learning rate of $0.00001$ with a weight decay of $0.01$ for the adam optimizer with weight decay \citep{loshchilov2018decoupled}. The LoRA configuration, if applicable, is $8$ for the rank of the update matrices, $32$ for the scaling factor, and $0.1$ for dropout. We are certain that the performance of our approach can be further improved if one focuses on a singular dataset and finds the optimal hyperparameter configuration for each dataset, but the computational cost of doing such a hyperparameter search is immense and beyond our financial scope and the general scope of this paper, which aims to point out the general merits of our approach.

All experiments were run on a shared GPU cluster outfitted with the 40GB and 80GB versions of the Nvidia A100 GPU, an AMD EPYC 7742 CPU, and 512GB of RAM. The code is implemented in PyTorch and PyTorch Lightning, and the initial model weights were loaded from HuggingFace.

\subsection{Data}

We train and test on a total of eight datasets to show where our approach demonstrates notable and promising performances. Special attention is places on the most prominent of these eight, the CoNLL--2003 dataset \citep{tjong-kim-sang-de-meulder-2003-introduction} sporting four different entity classes and its second iteration CoNLL++ \citep{wang-etal-2019-crossweigh}, which corrected 5.38\% of the apparently wrongly annotated test sentences. 

Furthermore, we include the OntoNotes \citep{pradhan-etal-2013-towards} and Few--NERD \citep{ding-etal-2021-nerd} datasets, which are similar to CoNLL--2003 but have more granular entities (18 and 66 entity classes, respectively). For example, whereas in CoNLL--2003, we only have a coarse-grained entity type ``Person'', this is split into eight types in Few-NERD: ''Actor``, ''Artist/Author``, ''Athlete``, ''Director``, ''Politician``, ''Scholar``, ''Soldier``, and ''Other``.

Going a different route, the WNUT-17 \citep{derczynski-etal-2017-results} dataset features six different entity classes and focuses on identifying unusual, previously-unseen entities in the context of emerging discussions. We also include three domain-specific datasets, two focusing on biomedical named entities (JNLPBA in \citet{collier-kim-2004-introduction} and NCBI-Disease in \citet{ncbi}) and one on financial ones (FiNER-ORD in \citet{shah2023finer}). The two bio-medical datasets have five and one different entity classes, respectively, and the financial NER dataset has three.

The combined length of this dataset is 290,317 sentences for the training set, 42,016 for the validation set, and 60,477 sentences for the test set.

\subsection{Coarse-tuning}

\begin{table}[t]
    \centering
    \begin{tabular}{lcccc}
    \toprule
    Model & Size & Dataset & LoRA & Micro-F$_1$ in  \%
    \\
    \midrule
    GPT2-XL & 1.5b & All & No & 72.91\\
    RedPajama & 3b & All & No & 73.61\\
    Falcon & 7b & All & Yes & 62.64\\
    Llama & 7b & All & Yes & 71.81\\
    Llama & 13b & All & Yes & 70.86\\
    \midrule
    GPT2-XL & 1.5b & Bio & No & 79.30\\
    BioMedLM & 2.7b & Bio & No & 81.31\\
    Llama & 7b & Bio & Yes & 76.18\\
    
    \bottomrule
    
    \end{tabular}
    \caption{Results of coarse-tuning various models on our combined Named Entity Decoding dataset. The LoRA column signals if a low-rank adaptation \citep{hu2021lora} was applied. The table is split into two, the first half reports performance when all datasets are combined and the second when we only consider the two biomedical datasets.  We train each model for 15 epochs and report the best micro-F$_1$ on the combined validation set. Due to its weak performance during this step, we do not continue our experiments with the Falcon model.}
    \label{tab:pretraining}
\end{table}

As a first step, we merge all training splits of the datasets discussed before and train a language model on the task of predicting the entity string $E$. We call this step ``coarse-tuning'' the pre-trained language model, as we infuse the model with a general sense of ``what named entities are''. We do not apply iNERD during the validation phase to simplify this step. The results are reported in Table~\ref{tab:pretraining}. 

It should be noted that the models have to deal with quite noisy data, as the entity type tokens $\xi$ are not the same among the datasets. Take the CoNLL-2003 and Few-NERD datasets for example. The former has four different entity type tokens, whereas the latter has 66. Nevertheless, we theorize that by simply letting the model get exposure to the general structure introduced in Equation~\ref{eq:nerd} it can gather valuable insights and might even understand the link between a coarse-grained entity type like ``Organization'' (in CoNLL-2003) and its fine-grained subtype ``Company'' (in Few-NERD).

\subsection{Results without dataset specific fine-tuning}

\begin{table*}[t]
    \centering
    \scriptsize
    \begin{tabular}{l@{\qquad}ccccccccc}
    \toprule
    \multirow{2}{*}{\raisebox{-\heavyrulewidth}{Model}}  & \multicolumn{8}{c}{Results after coarse-tuning in micro-F$_1$ in \% on dataset ... } \\
    \cmidrule{2-9}
    & CoNLL-2003 & CoNLL++ & OntoNotes & Few-NERD & WNUT-17 & JNLPBA & NCBI-Disease & FiNER-ORD
    \\
    \midrule
    \multicolumn{9}{l}{\textbf{iNERD + ...}}
    \\
    GPT2-XL & 89.57 & 90.54 & 83.39 & 50.95 & 43.40 & 58.49 &79.30 & 75.96\\
    BioMedLM & - & - & - & - & - & \textbf{59.06} & \textbf{84.05} & -\\
    RedPajama & \textbf{91.06} & \textbf{92.09} & \textbf{86.93} & \textbf{51.25} & \textbf{49.03} & 57.65 & 81.17 & \textbf{80.69} \\
    Llama-7b & 90.33 & 91.83 & 83.19 & 51.01 & 43.41 & 46.70 & 77.46 & 74.13\\
    Llama-13b & 90.88 & \textbf{92.09} & 81.68 & 50.22 & 39.90 & 57.85 & 75.37 & 75.56\\    
    \bottomrule
    
    \end{tabular}
    \caption{Micro-F$_1$ in \% on each dataset before fine-tuning and after coarse-tuning each model on the complete dataset, except BioMedLM, which was only coarse-tuned on the bio-medical domain. We applied LoRA to all models with a size above 3 billion parameters. The model sizes are as reported in Table~\ref{tab:pretraining}. We do not report the performance of bio-medical coarse-tuned GPT2-XL and Llama-7b variations, as they show worse performances than the general coarse-tuned ones.}
    \label{tab:zeroshotresults}
\end{table*}

Looking at Table~\ref{tab:zeroshotresults}, it becomes apparent that strong performances across datasets are attainable without applying specific fine-tuning on the respective training dataset. An interesting observation is that a larger model size does not consistently yield improved performance outcomes. Our largest studied model, the 13-billion parameter version of Llama, can mostly beat its smaller sister, the 7-billion version, but is largely overcome by the drastically smaller RedPajama (3-billion parameter). We theorize that the most likely explanation for this phenomenon is that during coarse-tuning, we apply LoRA to both Llama models to be able to train them in a reasonable time frame, which reduces the number of trainable parameters drastically. Therefore, for datasets with many entity classes $\xi$, like Few-NERD and OntoNotes, models with LoRA applied struggle to learn the subtle nuances between different classes and thus fail to outperform smaller models, likely because their available updateable parameter size is simply too small to fit these nuances. 

Another insight is that pre-training on a specific domain helps the model during named entity decoding immensely, as shown in the performance of the BioMedLM model. We see this as a vast opportunity for domain-specific pre-training of generative language models to make smaller models usable for the iNERD approach.

Even though the performances reported are \textit{not} zero-shot, as a small part of the coarse-tuning dataset consists of the training dataset of the respective dataset, this still demonstrates the impressive capabilities of such a model, the coarse-tuning routine, and the iNERD algorithm, as later shown in the ablation study.

\subsection{Fine-tuning results}

\begin{table*}[t]
    \centering
    \scriptsize
    \begin{tabular}{l@{\qquad}ccccccccc}
    \toprule
    \multirow{2}{*}{\raisebox{-\heavyrulewidth}{Model}}  & \multicolumn{8}{c}{Fine-tuning results in micro-F$_1$ in \% on dataset ... } \\
    \cmidrule{2-9}
    & CoNLL-2003 & CoNLL++ & OntoNotes & Few-NERD & WNUT-17 & JNLPBA & NCBI-Disease & FiNER-ORD

    \\
    \midrule
    \multicolumn{9}{l}{\textbf{iNERD + ...}}
    \\
    GPT2-XL & 91.51 & 92.71 & 86.15 & 51.63 & 53.25 & 58.70 &83.79 & 81.69\\
    BioMedLM & - & - & - & - & - & \textbf{60.08} & \textbf{86.37} & -\\
    RedPajama & 91.06 & 92.09 & \textbf{87.71} & \textbf{51.81} & 55.26 & 59.38 & 85.75 & 82.82\\
    Llama-7b & 92.75 & 94.10 & 84.27 & 51.72 & 55.59 & 57.91 & 80.81 & 82.42\\
    Llama-13b & \textbf{93.09} & \textbf{94.21} & 84.58 & 51.13 & \textbf{55.76} & 59.27 & 85.07 & \textbf{83.75}\\

    \midrule
    
    BERT-Base & 92.4 & - & - & - & - & - & 86.37 & -\\
    BERT-Large & 92.8  & - & - & - & - & - & - & -\\
    BioBERT & -  & - & - & - & - &  	\textbf{77.59}  & \textbf{89.71} & -\\
    PL-Marker & \textbf{94.0}  & - & \textbf{91.9} & \textbf{70.9} & - &  -  & - & -\\
    FiNER-LFs & - & - & - & - & - & - & - & \textbf{79.48}\\
    CrossWeigh & 93.43  &  94.28  & - & - & 50.03 & - & - & -\\
    CL-KL& 93.85  & \textbf{ 94.81}   & - & - &  	\textbf{60.45}  & - &88.96  & -\\
    
    \bottomrule
    
    \end{tabular}
    \caption{Micro-F$_1$ in \% on each test dataset after fine-tuning. The table is divided into two parts. The first shows the performance of iNERD plus a generative language model. The second part shows the performances of various encoder-only approaches. BERT-Base and BERT-Large are taken from \citet{devlin2018bert}, BioBERT from \citet{biobert}, PL-Marker from \citet{ye-etal-2022-packed}, FiNER-LFs from \citet{shah2023finer}, CrossWeigh from \citet{wang-etal-2019-crossweigh}, and CL-KL from \citet{wang-etal-2021-improving}.}
    \label{tab:finetuneresults}
\end{table*}

After evaluating the iNERD approach on its capabilities after coarse-tuning, we further fine-tune it on each dataset. The results of this can be seen in Table~\ref{tab:finetuneresults}. In there, we also report various competing approaches and their performances, taken from the respective papers. 

A first observation is that iNERD is capable of performing on par with or better than the standard encoder-only approach reported for the BERT \citep{devlin2018bert} model. A more general observation is that iNERD performs considerably well on datasets with a smaller entity class size, like CoNLL-2003 or NCBI-Disease. For our main focus, the datasets CoNLL-2003 and its corrected version CoNLL++, iNERD is able to be almost on par with competing state-of-the-art encoder-only approaches \citep{ye-etal-2022-packed, wang-etal-2019-crossweigh}, which are complex implementations and are thus in stark contrast to our simple and still effective approach.

On the one hand, it struggles especially on Few-NERD and OntoNotes, where the entity class size is significantly larger. Furthermore, the fine variations of various bio-medical terms in JNLPBA and the novel entities in WNUT-17 seem also to be a considerable hurdle for our approach. Of course, one could have simply excluded these datasets from this study, but we want to point out fields where our approach is struggling, where it might be improved upon with further research, and therefore, not simply ignore possible drawbacks of our method.

Nevertheless, on the other hand, we surpass the current best-performing model on FiNER-ORD, beating it by a considerable margin of more than 4\% F$_1$ and establishing a new state-of-the-art for financial named entity recognition on this dataset.

In total, the results of our approach are promising for the concept of using generative language models for tasks that they are not originally intended for, as we show that our relatively simple approach can surpass the comparatively simple one proposed in \citet{devlin2018bert}.

\subsection{Ablation study}

\begin{table}[t]
    \centering
    \begin{tabular}{l@{\qquad}cc}
    \toprule
    \multirow{2}{*}{\raisebox{-\heavyrulewidth}{Approach}}  & \multicolumn{2}{c}{Results in micro-F$_1$ in \%} \\
    \cmidrule{2-3}
    & no fine-tuning & fine-tuning \\
    \midrule
    \multicolumn{3}{c}{\footnotesize\textit{CoNLL-2003}}\\
    
    iNERD + Llama-7b & 90.33 & 92.75\\
    - informed decoding & 86.52 & 92.43 \\
    - coarse-tuning & 0.0 & 91.81 \\
    - both & 0.0 & 91.72 \\
    \midrule
    \multicolumn{3}{c}{\footnotesize\textit{CoNLL++}}\\
    
    iNERD + Llama-7b & 91.83& 94.10 \\
    - informed decoding & 87.81 & 93.71 \\
    - coarse-tuning & 0.0 & 93.14 \\
    - both & 0.0 & 93.01 \\
    
    \bottomrule
    
    \end{tabular}
    \caption{Reported here are the micro-F$_1$ scores in \% on the test set of CoNLL-2003 and CoNLL++ for the original iNERD approach with a Llama-7b language model and the scores when we subtract either the informed decoding algorithm (see Algorithm \ref{alg:iNERD}), the coarse-tuning step, or both.}
    \label{tab:ablation}
\end{table}

To show the advantages of each component of our approach, we conduct an ablation study on the CoNLL-2003 and CoNLL++ datasets. The results are shown in Table~\ref{tab:ablation}.

As seen there, each component of the iNERD approach adds to the overall performance. If we subtract the coarse-tuning as well as informed decoding steps, the micro-$F_1$ score falls to a paltry but expected $0$\% for the no fine-tuning environment, similarly when we only exclude the coarse-tuning step. Not so momentous, but still significant, the informed decoding described in Algorithm~\ref{alg:iNERD} adds around $4$\% improvement for both datasets.

A similar, but not so severe, picture can be observed during fine-tuning, where the distance between each step subtracted shrinks, but is still present. In such a setting, we observe an overall improvement of more than $1$\% for the CoNLL-2003 and CoNLL++ datasets when we compare the complete approach to the one with all components turned off.

\section{Conclusion}

We introduced a novel approach for named entity recognition (NER) which leverages the outstanding language understanding capabilities of modern large language models (LLMs). Our Informed Named Entity Recognition Decoding (iNERD) algorithm is easy to implement and arguably as simple as an ``encoder-only'' transformer plus multilayer-perceptron classifier approach as proposed in the seminal BERT \citep{devlin2018bert} paper. It builds on top of recent LLMs and is thus future-proof, as the employed LLMs can easily be replaced by improved models whenever they become available. It furthermore incorporates an informed decoding scheme which further improves performance, eliminates any risk of hallucinations, and significantly increases the adaptability. This informed scheme leverages the named entity decoding structure proposed herein to mask out disallowed tokens during the prediction phase.

Extensive experimental validation shows the performance of our framework to be mostly on par with competing ``encoder-only'' approaches, if not better. Experiments further reveal considerable and outstanding adaptive capabilities and show that iNERD can react to changes in the underlying data distribution without any additional fine-tuning. This contrasts said ``encoder-only'' approaches, which dominate the current NER landscape, as these have to be retrained whenever their set of entity classes changes.

An obvious next step is testing the largest generative language models, like the 70 billion parameter version of Llama, the 40 billion parameter version of Falcon, or even the 176 billion parameter version of Bloom. Using these could improve the performance of the complete iNERD setup on each dataset even further. On the other hand, training these huge model variations is extremely expensive and beyond our current computational capabilities. As already discussed in the Results section, applying LoRA, a method to freeze certain parts of the model to allow training large models, likely leads to a performance decrease. This is yet another interesting path to take for future research, as one could try pre-training large language models without this technique to improve the downstream performance further. Similarly, one could increase the size of the coarse-tuning dataset and include even more datasets. 

Another promising starting point for future research is investigating how the various highly specialized named entity recognition techniques developed for encoder-only models like PL-Marker \citep{ye-etal-2022-packed} or Co-Regularization \citep{zhou-chen-2021-learning} can be applied to generative language models and iNERD to improve the performance further.

Different information extraction tasks like relation extraction or event identification are also clear candidates for future research, which we plan to tackle in a similar manner as iNERD, as these tasks are ``rigid'' like NER and would thus profit from an informed approach like the one we propose.

From a more practical standpoint, we plan to implement the iNERD approach in various real-world applications in the world of Financial Auditing and Bio-Medicine, for the advantages of our approach are clear: highly effective on unseen data with a variable entity set $\xi$ (see Table~\ref{tab:zeroshotresults}) and easily upgradeable with the newest large language model.

\section*{Acknowledgments}

This research has been funded by the Federal Ministry of Education and Research of Germany and the state of North-Rhine Westphalia as part of the Lamarr-Institute for Machine Learning and Artificial Intelligence. 

We would like to thank our colleagues and friends Armin Berger, Kostadin Cvejoski, Leonhard David, Maren Pielka, and Rajkumar Ramamurthy for providing valuable feedback on this paper.

\bibliographystyle{plainnat}
\bibliography{references}

\end{document}